\documentclass[letterpaper]{article} 

\ifdefined\aaaianonymous
    \usepackage[submission]{aaai2026}  
\else
    \usepackage{aaai2026}              
\fi

\usepackage{times}  
\usepackage{helvet}  
\usepackage{courier}  
\usepackage[hyphens]{url}  
\usepackage{graphicx} 
\usepackage{natbib}  
\usepackage{caption} 
\usepackage{amsfonts} 
\usepackage{amsmath}
\usepackage{booktabs}
\usepackage{algorithm}
\usepackage{algorithmic}

\usepackage{newfloat}
\usepackage{listings}
\DeclareCaptionStyle{ruled}{labelfont=normalfont,labelsep=colon,strut=off} 
\floatstyle{ruled}
\newfloat{listing}{tb}{lst}{}
\floatname{listing}{Listing}

\ifdefined\aaaianonymous
    \title{Monocular Mesh Recovery and Body Measurement of Female Saanen Goats}
\else
    \title{Monocular Mesh Recovery and Body Measurement of Female Saanen Goats}
\fi

\author{
    Bo Jin\textsuperscript{\rm 1}, ShichaoZhao\textsuperscript{\rm 1}, Jin Lyu\textsuperscript{\rm 2}, Bin Zhang\textsuperscript{\rm 1}, Tao Yu\textsuperscript{\rm 3}, Liang An\textsuperscript{\rm 4,}\textsuperscript{*}, Yebin Liu\textsuperscript{\rm 4,}\textsuperscript{*}, Meili Wang\textsuperscript{\rm 1,5,6,7,}\textsuperscript{*}
}
\affiliations{
    \textsuperscript{\rm 1}College of Information Engineering, Northwest A\&F University

    \textsuperscript{\rm 2}Department of Electronic and Electrical Engineering, Southern University of Science and Technology

    \textsuperscript{\rm 3}Beijing National Research Center for Information Science and Technology, Tsinghua University 
    
    \textsuperscript{\rm 4}Department of Automation, Tsinghua University
        
    \textsuperscript{\rm 5}Key Laboratory of Agricultural Internet of Things, Ministry of Agriculture

    \textsuperscript{\rm 6}Shaanxi Key Laboratory of Agricultural Information Perception and Intelligent Service

    \textsuperscript{\rm 7}Shaanxi Engineering Research Center of Agricultural Information Intelligent Perception and Analysis
    
     wml@nwsuaf.edu.cn;\{liuyebin,anliang\}@mail.tsinghua.edu.cn

}

\usepackage{bibentry}
\begin{document}

\maketitle

\begin{abstract}
The lactation performance of Saanen dairy goats, renowned for their high milk yield, is intrinsically linked to their body size, making accurate 3D body measurement essential for assessing milk production potential, yet existing reconstruction methods lack goat-specific authentic 3D data. To address this limitation, we establish the FemaleSaanenGoat dataset containing synchronized eight-view RGBD videos of 55 female Saanen goats (6-18 months). Using multi-view DynamicFusion, we fuse noisy, non-rigid point cloud sequences into high-fidelity 3D scans, overcoming challenges from irregular surfaces and rapid movement. Based on these scans, we develop SaanenGoat, a parametric 3D shape model specifically designed for female Saanen goats. This model features a refined template with 41 skeletal joints and enhanced udder representation, registered with our scan data. A comprehensive shape space constructed from 48 goats enables precise representation of diverse individual variations. With the help of SaanenGoat model, we get high-precision 3D reconstruction from single-view RGBD input, and achieve automated measurement of six critical body dimensions: body length, height, chest width, chest girth, hip width, and hip height. Experimental results demonstrate the superior accuracy of our method in both 3D reconstruction and body measurement, presenting a novel paradigm for large-scale 3D vision applications in precision livestock farming.
\end{abstract}

\begin{links}
\link{Code,Data}{https://github.com/bojin-nwafu/Female-Saanen-Goats}
\end{links}

\section{Introduction}
Female Saanen goats, originating from Switzerland, are globally recognized as a premier dairy breed and serve as crucial parental stock for the genetic improvement of dairy goat populations worldwide~\cite{zhang2025whole}. A strong correlation exists between their morphometric characteristics and milk production, motivating current international research efforts towards modernized and precise management strategies for this breed~\cite{wang2022genome}. Three-dimensional vision technology enables precision management of dairy goats by capturing rich, high-dimensional data for accurate individual monitoring, precise phenotyping, and targeted genetic improvement programs.

\begin{figure}[!tb]
    \centering
    \includegraphics[width=\linewidth]{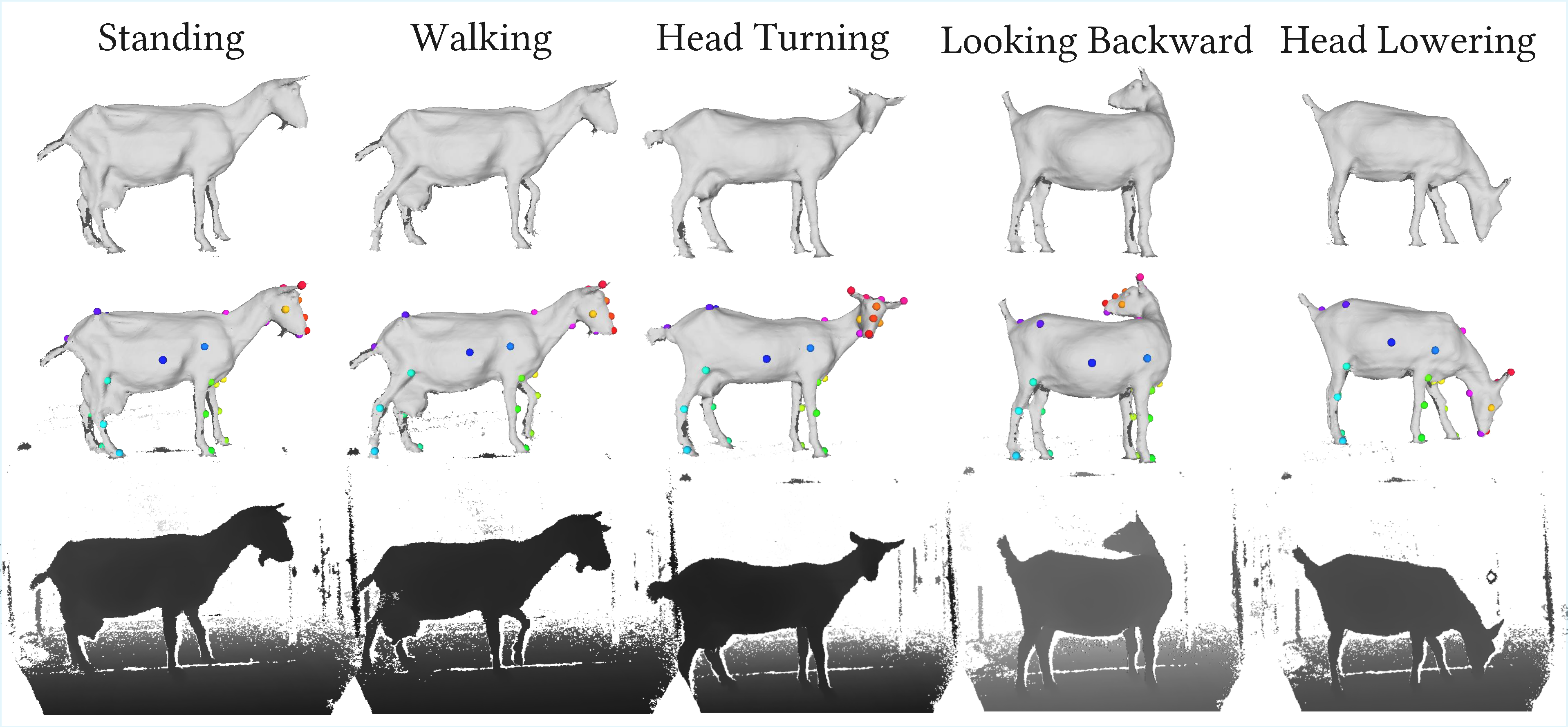}
    \caption{Pose variation examples in the Saanen dataset. (a) standing, (b) walking, (c) head turning, (d) looking backward, and (e) head lowering. Top row: fusion meshes after denoising; Middle row: manually labeled 3D surface keypoints; Bottom row: source depth images of one view.}
    \label{fig:pose_categories}
\end{figure}

However, the large-scale deployment of 3D vision technology in this domain faces substantial challenges. These primarily include the difficulty of acquiring high-quality animal data in dynamic, on-farm environments and the lack of robust parametric 3D models capable of accurately representing critical physiological structures, such as the udder. In conclusion, these factors hinder the widespread adoption and effective utilization of 3D vision technology in precision Saanen dairy goat farming. 

Previously, Zuffi et al.\cite{Zuffi_2017_CVPR} introduced the SMAL parametric model, which constructs a statistical model for quadrupeds using scans of 41 animal toys. However, this model lacks Saanen goat data and exhibits significant discrepancies with actual dairy goat physique in real farming environments, failing to adequately capture their morphometric characteristics. While Varen\cite{zuffi2024varen} demonstrated that high-precision species-specific parametric models can be constructed using high-quality laser scanning data, such systems face practical limitations: animals like sheep exhibit less docile behavior than horses, and the high system costs challenge deployment in farm environments. Hence, addressing the challenges of high-quality 3D data collection and precise parametric modeling for Saanen dairy goats remains crucial.
To address the scarcity of high-quality 3D data for Saanen dairy goats, we developed a lightweight eight-view RGBD camera array system for capturing synchronized video sequences in real farming environments. From this data collection effort, we constructed the FemaleSaanenGoat dataset.This dataset captures dynamic RGBD sequences of 55 female Saanen dairy goats aged 6 to 18 months, with each sequence recorded at 30 FPS for 30 seconds. The sequences encompass rich postures of goats in various natural states, including standing, walking, and head turning. We employed a multi-view DynamicFusion algorithm that effectively fuses noisy point cloud sequences to generate geometrically accurate and complete 3D scans. The dataset contains approximately 3,200 high-fidelity 3D scans (approximately 2,700 for training and 500 for testing), with each scan model manually annotated with 32 anatomical 3D keypoints to facilitate subsequent parametric model fitting.

Based on our new dataset,we develop the first real data-driven SaanenGoat parametric model incorporating anatomically relevant features, particularly mammary glands. The model uses pose-normalized 3D reconstructions from our multi-view system and enables high-precision 3D reconstruction from single-view RGBD input through learned shape priors and geometric constraints. Our approach automatically measures six critical body dimensions (body length, height, chest width/girth, hip width/height) via strategically defined keypoints, enabling precise morphometric analysis for livestock management. Experimental results demonstrate that our SaanenGoat model substantially outperforms the generic SMAL and SMAL+ models. Our model reduces 3D reconstruction errors by up to 77.7\% on the In-Shape test set and 66.5\% on the Out-Shape test set compared to SMAL, while maintaining strong generalization. For body measurement, our model achieves superior accuracy with MAE of 1.90, significantly outperforming SMAL (4.89) and SMAL+ (3.48).

In summary, our contributions are as follows:
\begin{itemize}
\item We introduce the first eight-view synchronized RGBD video dataset named FemaleSaanenGoat, which includes 55 individuals with different sizes and diverse poses.
\item We create the SaanenGoat parametric model by learning from the high-fidelity 3D scans and achieve precise 3D reconstruction.
\item We implement a automatic and accurate body size measurement system that leverages our parametric model and anatomically defined landmarks.
\end{itemize}

\begin{figure*}[htp]
    \centering
    \includegraphics[width=0.95\linewidth]{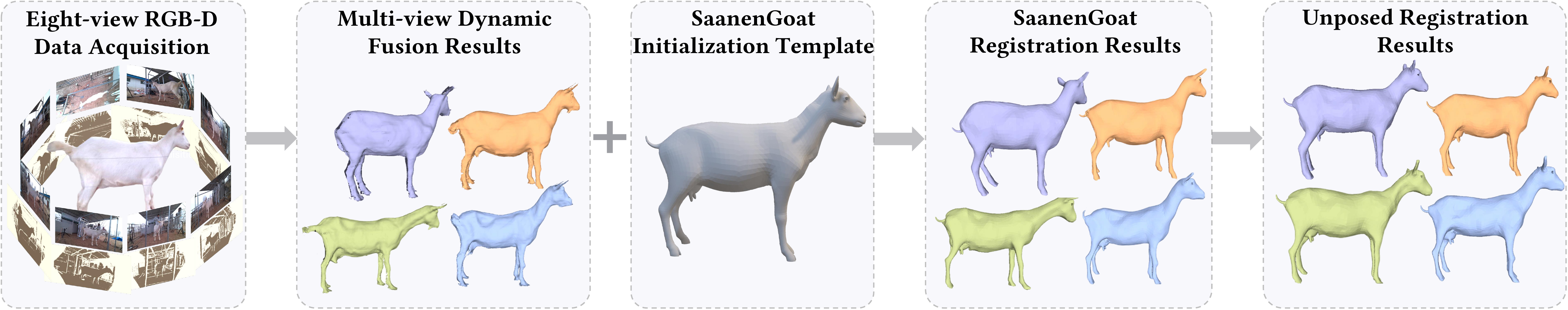}
    \caption{Data Processing Pipeline. We capture RGBD data of Saanen dairy goats using top-view and seven surrounding cameras. Multi-view dynamic fusion generates geometrically accurate 3D goat models. The Saanen Goat Initialization Template is registered to the scanned data, followed by pose normalization to T-pose for shape space construction.}
    \label{fig:workflow}
\end{figure*}


\section{Related Works}
\subsection{Quadrupedal Shape Model Creation}
We focus on parametric quadruped models. SMAL~\citep{Zuffi_2017_CVPR}, built from 41 toy animal scans, has limited shape space and cannot represent out-of-domain species like pigs, goats, or horses. Extensions to SMAL include vertex displacements for specific animals~\cite{zuffi2018lions} and scale parameters for dog sizing~\cite{biggs2020left}, but retain SMAL's bone structure and mean shape, limiting domain-specific representations.
Species-specific models include BARC~\cite{rueegg2022barc} and BITE~\cite{ruegg2023bite} (D-SMAL for dogs using breed information), hSMAL~\cite{li2021hsmal} (37 horse toy scans), and VAREN~\cite{zuffi2024varen} (anatomical skeletons with muscle-based deformations from 4k horse scans). AWOL~\cite{zuffi2024awol} proposes SMAL+ by integrating multiple animal models into a 145-dimension space. Other works target pigs~\cite{an2023three} and dogs~\cite{kearney2020rgbd} but lack generalization. No parametric models exist specifically for goats.
    
\subsection{Monocular Animal Mesh Recovery}

Monocular animal mesh recovery approaches fall into two categories: model-free and model-based methods. Model-free methods include explicit articulated reconstruction~\cite{wu2023magicpony,yao2022lassie,li2024learning} and implicit geometry prediction~\cite{hong2023lrm,sun2023dreamcraft3d,tochilkin2024triposr}. While effective for common pets, they struggle with rare agricultural animals.
Model-based methods rely on SMAL, limiting prediction to SMAL's shape space. Recent advances include optimization-based~\cite{Zuffi_2017_CVPR,zuffi2018lions,biggs2019creatures}, CNN-based~\cite{zuffi2019three,biggs2020left,rueegg2022barc,ruegg2023bite}, and Transformer-based~\cite{lyu2025animer} approaches. AniMer+ further extends this with a unified multi-taxa framework using synthetic dataset augmentation~\cite{lyu2025animer+}. Model capability and dataset content are crucial for recovery performance. While datasets like Animal3D~\cite{xu2023animal3d} and GenZoo~\cite{niewiadomski2024generative} support general pose and shape estimation, their mesh quality remains insufficient for applications requiring precision, such as body measurement.

\subsection{3D Body Measurement for Agriculture}
Traditional manual body measurements using meter sticks or metric tape are labor-intensive, time-consuming, subjective, and can cause livestock stress, leading to inconsistent results. Computer vision and 3D imaging techniques have revolutionized this process. ~\cite{ruchay2020accurate} uses three-view depth cameras to obtain complete cow point clouds, localizing keypoints to calculate nine body measurements. ~\cite{du2022automatic} projects RGB-detected keypoints onto livestock point clouds, combining interpolation and position normalization for cow and pig measurements. \cite{hao2023improved} proposes an improved PointNet++ model segmenting pig point clouds into ten parts, localizing keypoints, and calculating parameters via least squares and slicing. Similar approaches apply to goats\cite{jin2024pointstack}. These methods input unordered point clouds. Alternatively, ~\cite{lu2025automatic} proposes a two-stage coarse-to-fine cattle measurement strategy, initially fitting 2D keypoints using SMAL, then refining surface meshes via Graph Convolutional Networks (GCN) for key dimension estimation.

\section{Method}
\subsection{FemaleSaanenGoat Dataset}
\label{sec:sec:dataset}
\paragraph{System Setup.} 
We build a corridor-style acquisition setup with eight viewpoints (one top view and seven surrounding views) to capture high-quality 3D point clouds of goats, as shown in the supplementary material.The apparatus comprises a walking corridor with fencing, entry and exit gates, and photoelectric sensors that detect a goat’s arrival. 
We use eight synchronized Microsoft Azure Kinect DK RGBD cameras, capturing 1280x720 RGB videos and 640x480 depth videos at 30FPS. For each goat, we record 30-second videos. 

\begin{figure}[!tb]
    \centering
    \includegraphics[width=0.95\linewidth]{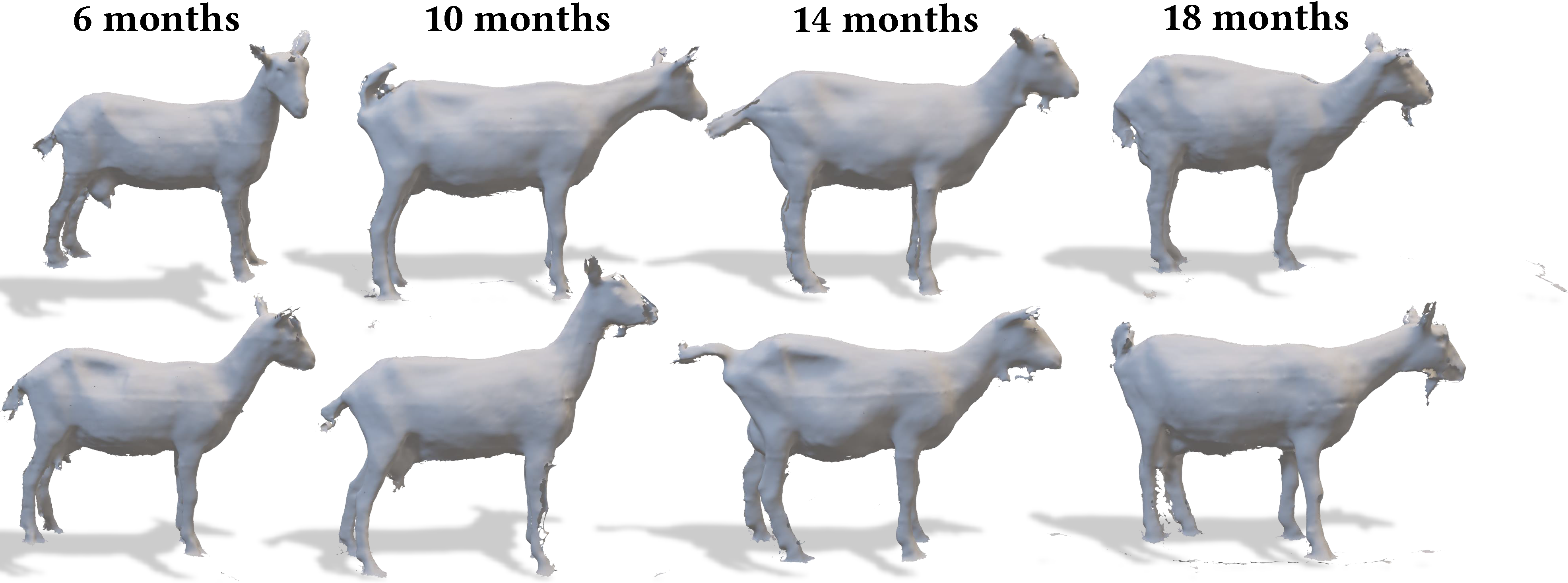}
    \caption{Examples of scan results for Saanen dairy goats of different ages and body types included in the dataset.}
    \label{fig:diversity_of_shapes.}
\end{figure}

\paragraph{Goat 3D Mesh Reconstruction.} We obtain initial camera poses through chessboard-based PnP calibration~\cite{zhang1999flexible}, then refine them using a temporally-constrained ICP optimization framework to achieve accurate 3D goat reconstruction. Given a RGB video, we use Track Anything~\cite{yang2023track} to obtain the foreground mask. To obtain smooth and coherent 3D meshes of deforming goat surfaces while accurately reconstructing the actual shape of goats, 
we modify DynamicFusion~\cite{newcombe2015dynamicfusion} to 8-view cases. 
As result, the reconstructed mesh exhibits low-noise surfaces and physically plausible deformations, accurately capturing the goat's shape and deformation over time. Additionally, we captured multiple poses in their natural state, as demonstrated in Fig.~\ref{fig:pose_categories}.
Afterwards, we apply radius filtering to the output meshes to further suppress noise, resulting in a smoother surface. Consequently, we collect a dataset named FemaleSaanenGoat that consists of 55 goats with ages ranging from 6 months to 18 months(Fig.~\ref{fig:diversity_of_shapes.}), which includes 48 goats with approximately 2,700 scans for model training, and 7 goats with around 500 scans for testing. The dataset contains synchronized eight-view RGBD video data, 2D keypoints, and corresponding three-dimensional mesh data for each frame. 

\subsection{Saanen Dairy Goat Parametric Model}
\label{sec:sec:SaanenGoat}
\paragraph{Initial Template. }
Despite the widespread adoption of Skinned Multi-Animal Linear (SMAL) models for quadruped representation, SMAL fails to accurately capture the Saanen dairy goat due to the absence of Saanen dairy goat samples in the 41 training toy scans. Therefore, this paper proposes a novel parametric model specifically constructed from our scan data of Saanen dairy goats, enabling accurate morphological representation.
\begin{figure}[!tb]
    \centering
    \includegraphics[width=0.85\linewidth]{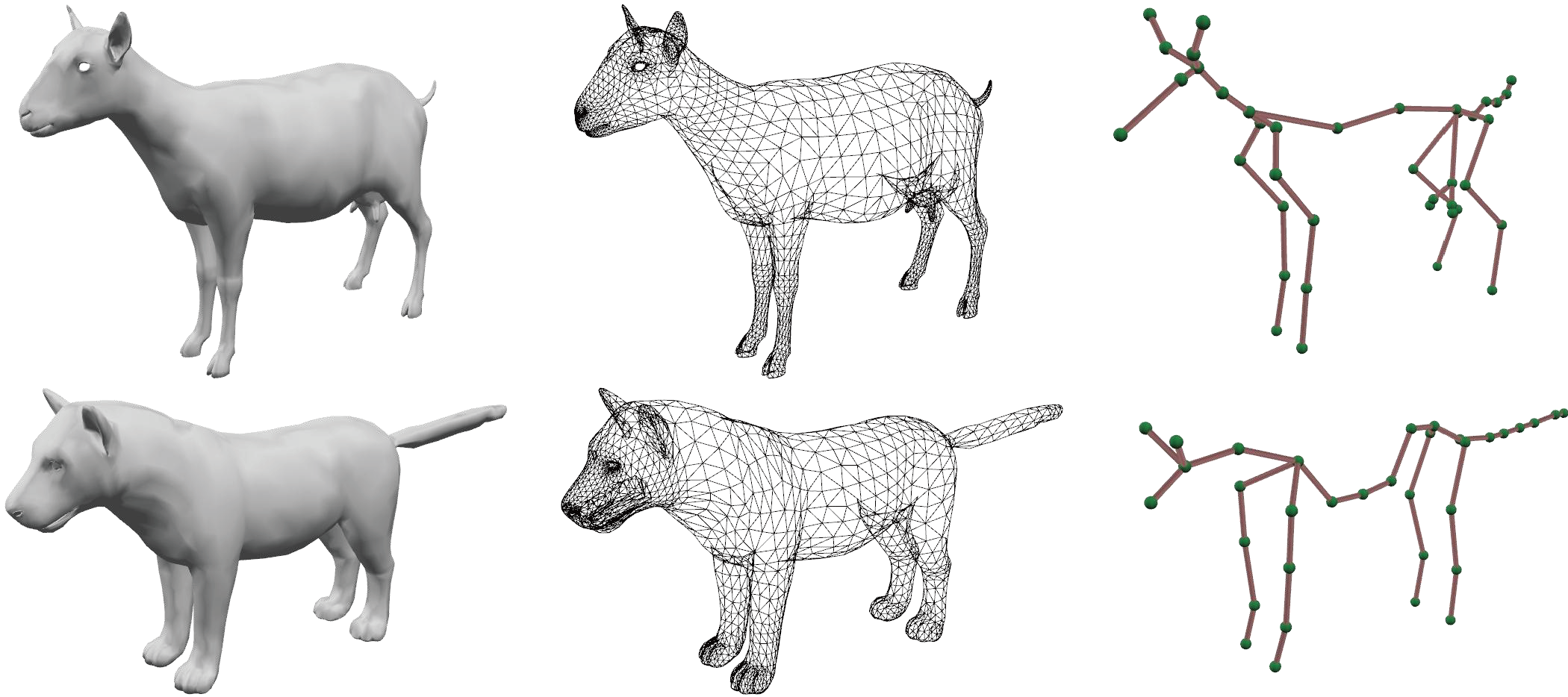}
    \caption{Comparison between the average model of our SaanenGoat (top) and the SMAL (bottom). Learned from real goat scan meshes, the SaanenGoat model provides a more accurate fit to the real goat's anatomy than SMAL.}
    \label{fig:goat_render}
\end{figure}
Specifically, we purchase a Saanen dairy goat model~\footnote{https://free3d.com/3d-model/goat-saanen-breed-rigged-5393.html} with skinning and skeletal binding as our initial template, comprising 39 skeletal joints, 3647 vertices and 7376 faces. However, the template's resolution inadequately captures fine-grained morphological features. We refine this template through two stages: (1) adjusting skeletal joint positions for anatomical accuracy according to caprine morphology and augmenting the udder region with additional joints for precise mammary structure control; (2) applying loop subdivision to increase mesh resolution and enhance morphological feature representation. 
The final template consists of a vertex set $\mathbf{V} \in \mathbb{R}^{N \times 3}$ containing $N=13,815$ vertices, a joint set $\mathbf{J} \in \mathbb{R}^{M \times 3}$ comprising $M=41$ joints, a skinning weight matrix $\mathbf{W} \in \mathbb{R}^{N \times M}$ mapping vertices to joints, and a kinematic tree defined by skeletal parent indices $\mathcal{P} = \{p_1, p_2, ..., p_M\}$, where $p_i$ denotes the parent joint of joint $i$. Fig.~\ref{fig:goat_render} compares our template with SMAL template. 

\paragraph{Pose Fitting to Scan.} 
We establish initial correspondence by defining 32 anatomically-informed landmarks on each scanned mesh (Fig.~\ref{fig:pose_categories}), strategically distributed across the head, neck, limbs, thorax, abdomen, pelvis, and tail for comprehensive morphological coverage. After scale normalization to establish uniform scaling, we compute optimal rigid transformation parameters (rotation matrix R and translation vector t) that minimize distances between corresponding landmarks. This rigid alignment provides coarse registration, establishing preliminary spatial correspondence for subsequent fine-scale deformation analysis.
\begin{equation}
\begin{aligned}
\{R, t\} = \arg\min_{R,t} \sum_{i=1}^n ||R \cdot v_i^S + t - v_i^T||^2
\end{aligned}
\end{equation}
where $v_i^S$ and $v_i^T$ represent corresponding points on the scanned mesh and template mesh, respectively. This transformation is solved by Singular Value Decomposition (SVD).

We optimize skeletal pose using these landmarks, with pose parameters in axis-angle format converted to rotation matrices via Rodrigues' formula. Global joint transformations are computed recursively through the kinematic tree, and Linear Blend Skinning (LBS) calculates vertices under pose transformation. We formulate pose parameter optimization through a joint loss function:

\begin{equation}
\begin{aligned}
\mathcal{L}_{\text{total}} &= \lambda_1\mathcal{L}_{\text{scale}} + \lambda_2\mathcal{L}_{\text{landmark}} + \lambda_3\mathcal{L}_{\text{correspondence}} \\ 
&+ \lambda_4\mathcal{L}_{\text{collision}} + \lambda_5\mathcal{L}_{\text{pose\_reg}},
\end{aligned}
\end{equation}
where \(\mathcal{L}_{\text{scale}}\) ensures that the two meshes are consistent in scale, avoiding mismatches caused by scale differences, \(\mathcal{L}_{\text{landmark}}\) ensures that the keypoints of the registration results align with the annotated keypoints, \(\mathcal{L}_{\text{correspondence}}\) ensures that the registration results align with the scans. Additionally, \(\mathcal{L}_{\text{collision}}\) represents the collision loss term, which prevents interpenetration artifacts between the udder and limbs of the dairy goat in various poses, and \(\mathcal{L}_{\text{pose\_reg}}\) prevents the registration results from exhibiting ill-poses. In our experiments, the values of \(\lambda_1, \lambda_2, \lambda_3\, \lambda_4\), and \(\lambda_5\) are set to 0.1, 0.1, 0.2, 0.1, and 0.15, respectively.

\paragraph{Shape Fitting to Scan.} Similar to pose optimization, we enhance fitting accuracy through mesh refinement. We compute a vertex displacement field that deforms the pose-aligned template mesh to better approximate the target geometry. In details, given the posed mesh $M_{pose}$ and the target mesh $M_{target}$, our objective is to determine an optimal vertex displacement field $\Delta V$ such that the deformed mesh $M_{deformed} = M_{pose} + \Delta V$  closely approximates the target while preserving mesh topology and surface smoothness.
Due to the inconsistent vertex density across different regions of the model, we partition the optimization process into three anatomical regions—torso, head, and limbs—to capture the morphological characteristics of the dairy goat with high fidelity. The optimization objective is as follows:
\begin{equation}
\begin{aligned}
L_{total} = \sum_{part \in \{body, head, legs, other\}} L_{part}+\lambda L_{global},
\end{aligned}
\end{equation}
where \(\lambda\) = 0.2, and the loss function  $L_{part}$ for each anatomical region comprises the following components.
\begin{equation}
\begin{aligned}
L_{part} &= w_{vc} \cdot L_{vertex} + w_{s} \cdot L_{smooth} \\
&+ w_{e} \cdot L_{edge} + w_{n} \cdot L_{normal}
\end{aligned}
\end{equation}
where the loss function $L_{vertex}$  ensures that deformed vertices closely match their target mesh counterparts, $L_{smooth}$ preserves surface smoothness, $ L_{normal}$ maintains consistency among neighboring face normals, and $L_{edge}$ stabilizes edge lengths throughout the deformation process. In our experiments, the values of $L_{vertex}$ , $L_{smooth}$, $ L_{normal}$, and $L_{edge}$  are set to 0.5, 0.1, 0.1, and 0.1, respectively.

\paragraph{Shape Space.} Based on data obtained from previous pose and shape fitting processes, we select fitting results from 48 Saanen dairy goats to construct the shape space of our model. We first utilize the estimated pose parameters $\boldsymbol{\theta}$ to build the global transformation matrices $\mathbf{G}_j(\boldsymbol{\theta})$ for each joint and their corresponding inverse matrices $\mathbf{G}_j^{-1}(\boldsymbol{\theta})$. Subsequently, we employ inverse linear blend skinning (inverse LBS) to restore the dairy goat models from various poses back to the standard T-pose, thereby preserving the geometric characteristics and topological structure of the models. 

\begin{figure}[!b]
    \centering
    \includegraphics[width=0.8\linewidth]{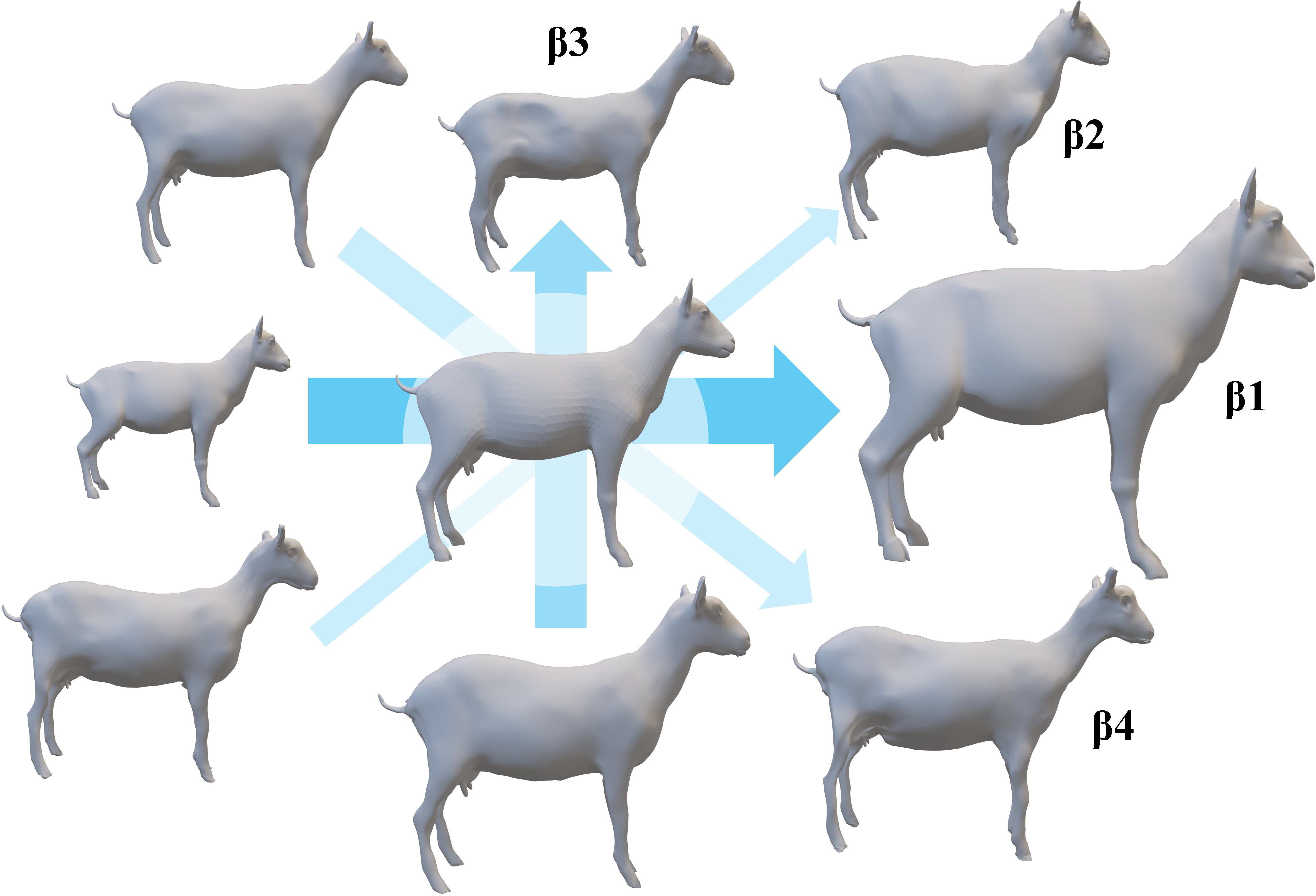}
    \caption{Visualization of the shape space for Saanen goats. The central grid represents the average shape, while the surrounding variations demonstrate different shape changes in the model. We visualize the first four principal components with deviations of ±2 standard deviations.}
    \label{fig:shape space}
\end{figure}

After normalizing all dairy goat models to the T-pose, we construct Shape Blend Shapes to capture shape variations across different individuals. Using the standard template mesh $\bar{\mathbf{T}}$ as the reference shape, we calculate the shape deviation for each normalized dairy goat model $\mathbf{D}^m$ as:
\begin{equation}
\begin{aligned}
\Delta\mathbf{D}^m = \mathbf{D}^m - \bar{\mathbf{T}}.
\end{aligned}
\end{equation}

We then compile all shape deviations into a shape deviation matrix $\tilde{\mathbf{D}}$ and perform Principal Component Analysis (PCA) on this matrix to obtain a set of orthogonal shape basis vectors and their corresponding eigenvalues. 
Based on the extracted shape basis vectors  $\mathbf{S}_i$, we define the Shape Blend Shapes function $\mathbf{BS}(\boldsymbol{\beta})$:
\begin{equation}
\begin{aligned}
\mathbf{BS}(\boldsymbol{\beta}) = \sum_{i=1}^{K}\beta_i\mathbf{S}_i,
\end{aligned}
\end{equation}
where $\boldsymbol{\beta} = [\beta_1, \beta_2, ..., \beta_K]^T$ is the shape parameter vector controlling the weights of each shape basis vector and \(K=48\). Each shape basis vector $\mathbf{S}_i$ has a dimension of $3N$, corresponding to the displacements of all vertices in the model.

Combining the standard template and the Shape Blend Shapes function, we construct a parameterized model for dairy goats. Given pose parameter \(\boldsymbol{\theta}\) and shape parameter \(\boldsymbol{\beta}\), the mesh generation process can be expressed as follows:
\begin{equation}
    \begin{split}
        \mathbf{v}_{s} &= \mathbf{v}_{t} + B\boldsymbol{\beta^{T}}, \\
        \mathbf{v} &= \mathrm{LBS}(\mathbf{v}_{s}, \boldsymbol{\theta}; W, J_{r}),
    \end{split}
\end{equation}
where \(\mathbf{v}_{t}\) is the template vertices, \(B\) is the shape blend shape weight, \(J_r\) is the joint regressor. This parametric representation enables efficient generation of three-dimensional dairy goat models with various body types by adjusting the low-dimensional shape parameters $\boldsymbol{\beta}$.

Following the SMPL ~\cite{loper2023smpl} framework, we implemented pose blend shapes but observed negligible improvements for Saanen dairy goats. As our body measurement analysis primarily concerns shape variations, we exclude pose blend shape demonstrations from this study. Furthermore, to preserve accurate morphological representation, we refrain from enforcing vertex symmetry operations due to inherent anatomical asymmetries in dairy goats, particularly in the lungs and mammary glands.

\subsection{Monocular Mesh Recovery}
\label{sec:sec:SaanenGoatSmalify}
To ultimately achieve fully automated monocular Saanen goat mesh recovery and body measurement, we extend the SMALify framework~\footnote{https://github.com/benjiebob/SMALify} – an efficient and widely-adopted tool for animal mesh optimization.
However, its generic formulation presents two critical limitations for our target domain: (1) The SMAL template lacks species-specific priors for goats, and (2) the optimization relies solely on 2D keypoints and silhouette constraints, lacking depth information and thus unable to capture true spatial scale. To address these challenges, we replace the SMAL model with our SaanenGoat model in SMALify and introduce two key innovations:
\begin{itemize}
    \item \textbf{3D Geometric Constraints}: We incorporate monocular depth constraints and contour-normal constraints during optimization, providing crucial 3D geometric cues that resolve inherent ambiguities in monocular reconstruction.
    \item \textbf{Biomechanical Regularization}: Drawing from the Saanen goat template, we enforce geometric constraints on joint angles and local deformations, preventing physiologically implausible mesh distortions.
\end{itemize}
The main loss function is given by:
\begin{equation}
\begin{aligned}
\mathcal{L_{\text{total}}} &= \underbrace{\lambda_{\text{sil}}\mathcal{L}_{\text{sil}}+ \lambda_{\text{kp}}\mathcal{L}_{\text{kp}}}_{\text{2D alignment}} + 
\underbrace{\lambda_{\text{depth}}\mathcal{L}_{\text{depth}}+\lambda_{\text{normal}}\mathcal{L}_{\text{normal}}}_{\text{3D alignment}} \\
& + \underbrace{\lambda_{\text{reg}}\mathcal{L}_{\text{reg}}}_{\text{Regularization}},
\end{aligned}
\end{equation}
where ${\lambda_{\text{sil}} = 0.1}$, ${\lambda_{\text{kp}} = 0.02}$, ${\lambda_{\text{depth}} = 0.002}$, ${\lambda_{\text{normal}} = 0.002}$, and ${\lambda_{\text{reg}} = 0.01}$. The loss formulation incorporates multi-modal constraints for robust 3D reconstruction: $\mathcal{L}_{\text{sil}}$ enforces 2D silhouette alignment between rendered contours and ground truth segmentation masks, ensuring projective consistency. $\mathcal{L}_{\text{kp}}$ measures the 2D keypoint reprojection error by comparing predicted and gt joint positions. For 3D geometric consistency, $\mathcal{L}_{\text{depth}}$ penalizes deviations between rendered depth maps and sensor measurements to resolve pose ambiguities, while $\mathcal{L}_{\text{normal}}$ regularizes surface deformation by matching predicted and observed surface normals, particularly along edge contours. $\mathcal{L}_{\text{reg}}$ serves as a regularization term that enforces biomechanical constraints to prevent anatomically implausible deformations.

\section{Experiments}


\subsection{Comparison with SMAL for Scan Registration}
\noindent\textbf{Datasets.} The evaluation of our parametric Saanen dairy goat model comprised two test sets. The In-Shape testset consists of five different pose categories (see \ref{fig:pose_categories}): standing, walking, head turning, looking backward, and head lowering. For each pose category, one goat is randomly selected, with each pose sequence containing 60 frames, totaling 300 frames in the test set. The Out-Shape test set contains 350 frames from 7 goats, that are not used in the construction of the shape blend shape. 

\noindent\textbf{Metrics.} In our experiments, we select the mean chamfer distance error and the mesh-to-scan distance error for vertices across all body parts as evaluation metrics following~\cite{zuffi2024varen}.

\noindent\textbf{Baselines.} Given the lack of parametric models specifically developed for Saanen dairy goats or other goat breeds in existing literature, this study adopts the original SMAL neutral template as the primary baseline framework. To provide a comprehensive evaluation, we additionally employ SMAL+~\cite{zuffi2024awol}, an extended version that encompasses a broader range of animal species, as a secondary comparative baseline.

\begin{figure}[t]     
    \centering     
    \includegraphics[width=0.95\linewidth]{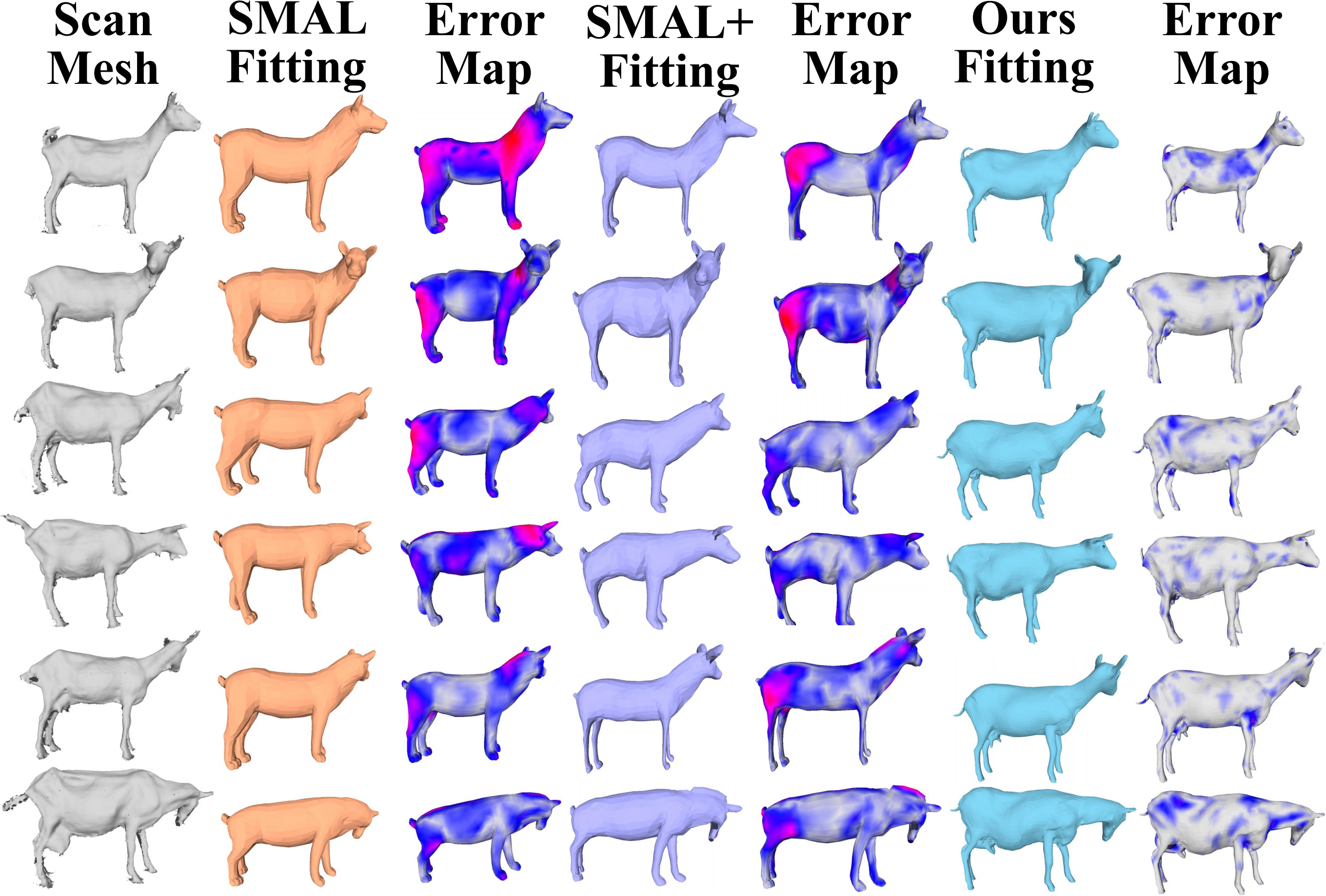}     
    \caption{
    Comparison of model fitting performance on scanned goat data: ground truth mesh, SMAL, SMAL+, and SaanenGoat fittings with corresponding distance error maps. Distance visualizations use blue-to-red gradients indicating minimal-to-maximal fitting errors.} 
    \label{fig:fit results} 
\end{figure}

Experiments were performed using a single NVIDIA GeForce RTX 3090 graphics card. We compare SMAL, SMAL+, and our SaanenGoat model on both datasets. Table~\ref{tab:in_shape_fitting_results} shows that SMAL exhibits higher fitting errors due to limited template diversity, while SMAL+ improves accuracy through expanded shape space from multiple species. Our SaanenGoat model achieves the lowest error rates by accurately capturing authentic Saanen goat body structure. Table~\ref{tab:out_shape_fitting_results} demonstrates our model maintains high accuracy on unseen subjects. Fig.~\ref{fig:fit results} visualizes results from the Out-Shape dataset, showing accurate 3D shape representation across various postures. Both quantitative and qualitative results confirm our model's generalization capability to unseen Saanen goats.

\begin{table}
\centering
\begin{tabular}{l cccc}
\toprule
& \multicolumn{2}{c}{Chamfer Distance} & \multicolumn{2}{c}{Mesh-to-Scan} \\
\cline{2-3} \cline{4-5}
& mean & std & Mean & std \\
\midrule
SMAL & 36.72 & 4.76 & 31.53 & 3.31 \\
SMAL+ & 31.18 & 3.26 & 25.14 & 2.79 \\
Ours & 10.65 & 0.62 & 7.02 & 0.46 \\
\bottomrule
\end{tabular}
\caption{Fitting Results on the In-Shape testset. Compared to both the neutral template SMAL model and SMAL+ model. Errors in mm.}
\label{tab:in_shape_fitting_results}
\end{table}

\begin{table}[bp]
\centering
\begin{tabular}{l cccc}
\toprule
& \multicolumn{2}{c}{Chamfer Distance} & \multicolumn{2}{c}{Mesh-to-Scan} \\
\cline{2-3} \cline{4-5}
& mean & std & Mean & std \\
\midrule
SMAL & 39.68 & 5.14 & 33.71 & 4.35 \\
SMAL+ & 32.46 & 3.75 & 26.32 & 2.87 \\
Ours & 15.30 & 1.83 & 11.29 & 1.76 \\
\bottomrule
\end{tabular}
\caption{
Out-Shape testset results. Comparison of reconstruction errors between our proposed model and the SMAL and SMAL+ baseline models. Errors in mm.
}
\label{tab:out_shape_fitting_results}
\end{table}

\subsection{Body Measurement Accuracy}
\noindent\textbf{Datasets} We collected body measurements from 20 Saanen dairy goats with varying ages and conformations for validation. Manual measurements served as ground truth for body dimensions. Previous research on Saanen dairy goat measurement~\cite{jin2024pointstack} used three-perspective Kinect sensors to acquire point clouds and calculated body dimensions through keypoint-based segmentation. After acquiring this dataset, we performed comparative evaluation against these measurements.

\noindent\textbf{Metrics} Mean Absolute Error (MAE) and Mean Absolute Percentage Error (MAPE) were selected as evaluation metrics. MAE represents the average of the absolute differences between predicted and actual values, providing a direct measure of prediction accuracy in the original units. MAPE, calculated as the average of the absolute percentage errors relative to the actual values, offers a scale-independent metric that facilitates comparison across different measurement scales and body size.

\begin{figure}[!tb]
    \centering
    \includegraphics[width=0.95\linewidth]{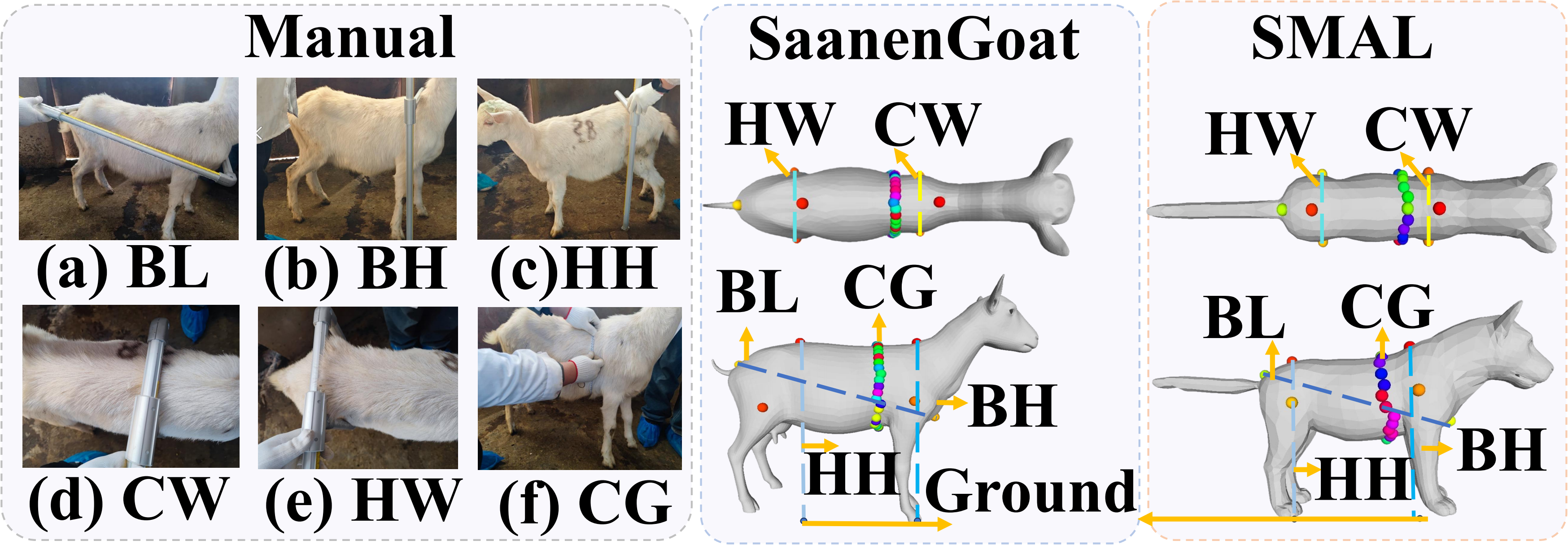}
    \caption{The methodology for body size measurement.BL: Body Length, BH: Body Height, HH: Hip Height, CW: Chest Width, HW: Hip Width, CG: Chest Girth.  Left: manual measurement. Middle: key point-based measurements derived from the SaanenGoat model. Right: key point-based measurements derived from the SMAL model.}
    \label{fig:body measurements definition compare}
\end{figure}

\begin{figure}[!tb]
    \centering
    \includegraphics[width=0.9\linewidth]{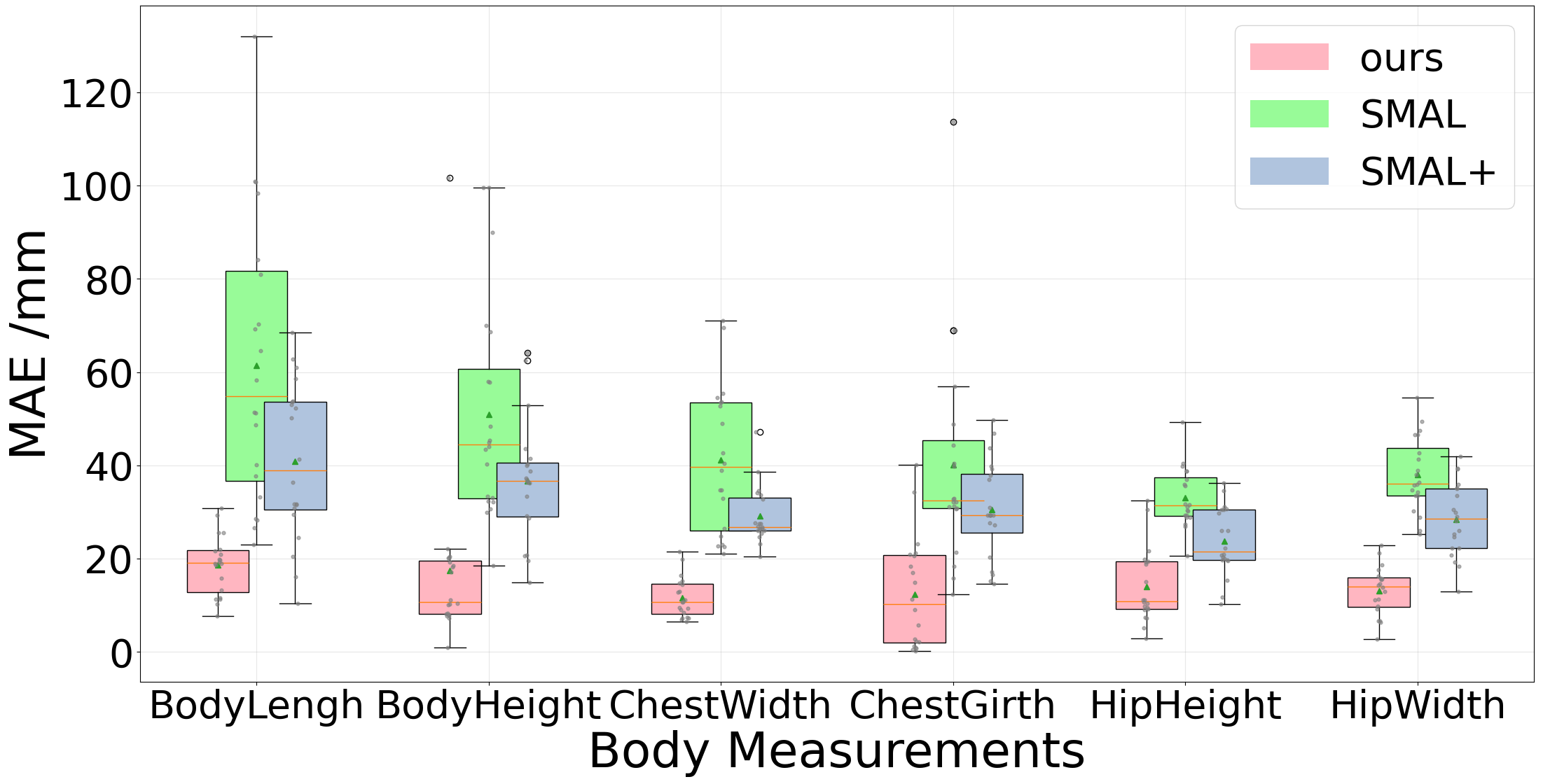}
    \caption{Body measurement comparison between SMAL, SMAL+, and our SaanenGoat model. Our model achieves closer agreement with manual measurements by more accurately capturing dairy goat morphology.
    }
    \label{fig:comparison_results}
\end{figure}

This paper measures six primary parameters: body length, body height, hip height, chest width, hip width, and chest girth, with measurement locations shown in Fig.~\ref{fig:body measurements definition compare}. We established corresponding morphometric keypoints on our parametric model based on anatomical expertise. Body length, chest width, and hip width are computed using Euclidean distances between keypoint pairs, while body height and hip height are calculated as distances from keypoints to the ground plane. For chest girth, we project the non-coplanar chest keypoints onto a reference plane, fit an ellipse to their distribution, and compute the elliptical perimeter. Fig.~\ref{fig:comparison_results} shows MAE error analysis comparing body size parameters from SaanenGoat, SMAL, and SMAL+ models against ground truth data. Results demonstrate that our SaanenGoat model achieves superior performance across all six body measurement parameters, with significant error reductions of 41-72\% compared to SMAL and SMAL+. The results of MAPE are presented in the supplementary material. 

Jin et al.\cite{jin2024pointstack} employed PointStack networks for point cloud segmentation, localized measurement keypoints from anatomical regions, and calculated corresponding metrics. Using their point cloud data, we fit our parametric model to compute body dimensions, with visualization results presented in the supplementary material.Table\ref{tab:body_size_comparison} shows Our model outperforms PointStack in body size prediction across most metrics. For MAE, we achieve lower errors in all 6 body parts, with the largest improvement in chest girth (45.6\% reduction). For MAPE, we outperform on 5 of 6 metrics, with notable improvements in chest girth (46.7\%) and Hip Width (26.0\%). Overall, our approach demonstrates superior accuracy and stability compared to point cloud segmentation methods.

\begin{table}
\centering
\begin{tabular}{lcccc}
\toprule
& \multicolumn{2}{c}{Ours} & \multicolumn{2}{c}{PointStack} \\
Body Size & MAE & MAPE & MAE & MAPE \\
\midrule
Body Length & 21.9 & 2.81 & 25.7 & 3.24 \\
Body Height & 16.8 & 2.35 & 18.3 & 2.54 \\
Chest Width & 14.6 & 6.81 & 11.3 & 5.43 \\
Chest Girth & 15.5 & 1.64 & 28.5 & 3.08 \\
Hip Height & 13.1 & 1.77 & 16.2 & 2.16 \\
Hip Width & 13.4 & 5.79 & 10.2 & 4.59 \\
\bottomrule
\end{tabular}
\caption{
Comparison with improved PointStack-based body dimension measurements derived from tri-view point cloud data. Errors in mm.
}
\label{tab:body_size_comparison}
\end{table}

\subsection{Monocular Mesh Recovery Accuracy}
\begin{figure}[ht]
    \centering
    \includegraphics[width=0.95\linewidth]{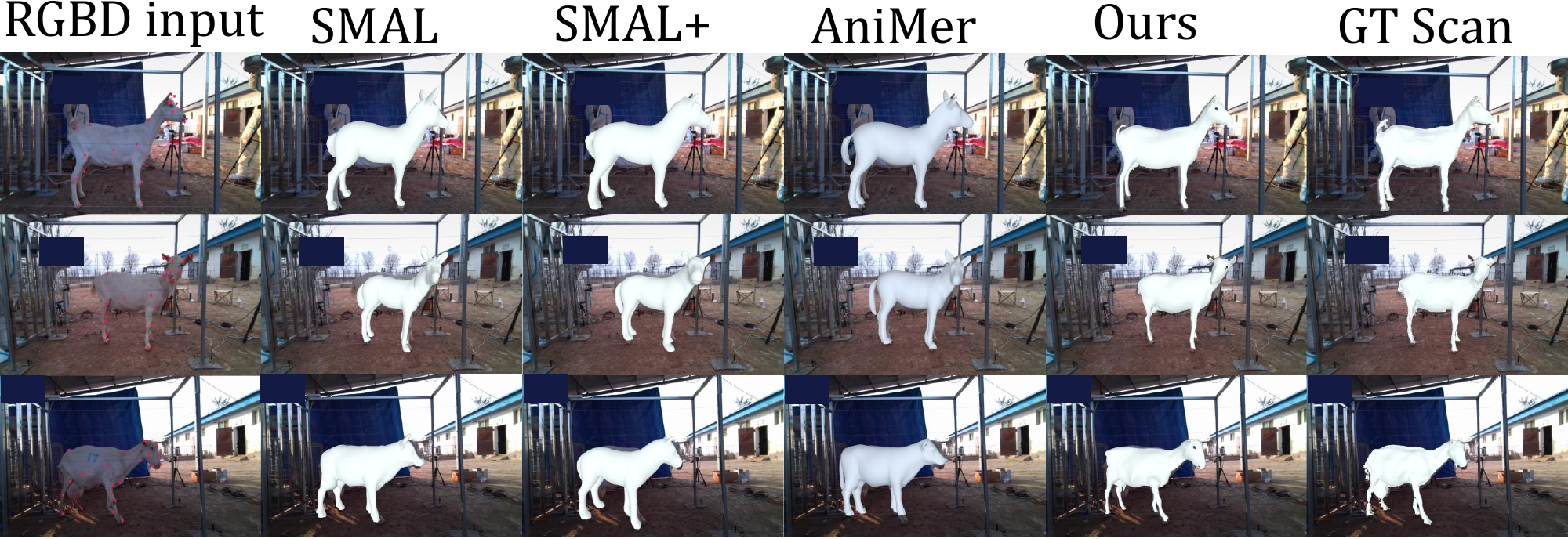}
    \caption{Monocular 3D goat mesh recovery results. From left to right:(a) Input RGBD image; (b) SMALify with SMAL model; (c) SMALify with SMAL+ model; (d) AniMer model's predicted mesh overlaps with the input data; (e) Our model's predicted mesh overlaps with the input data;  (f) Ground truth (GT) mesh overlaps with the RGB image. }
    \label{fig:monocular_results}
\end{figure}

\begin{figure}[ht]
    \centering
    \includegraphics[width=0.87\linewidth]{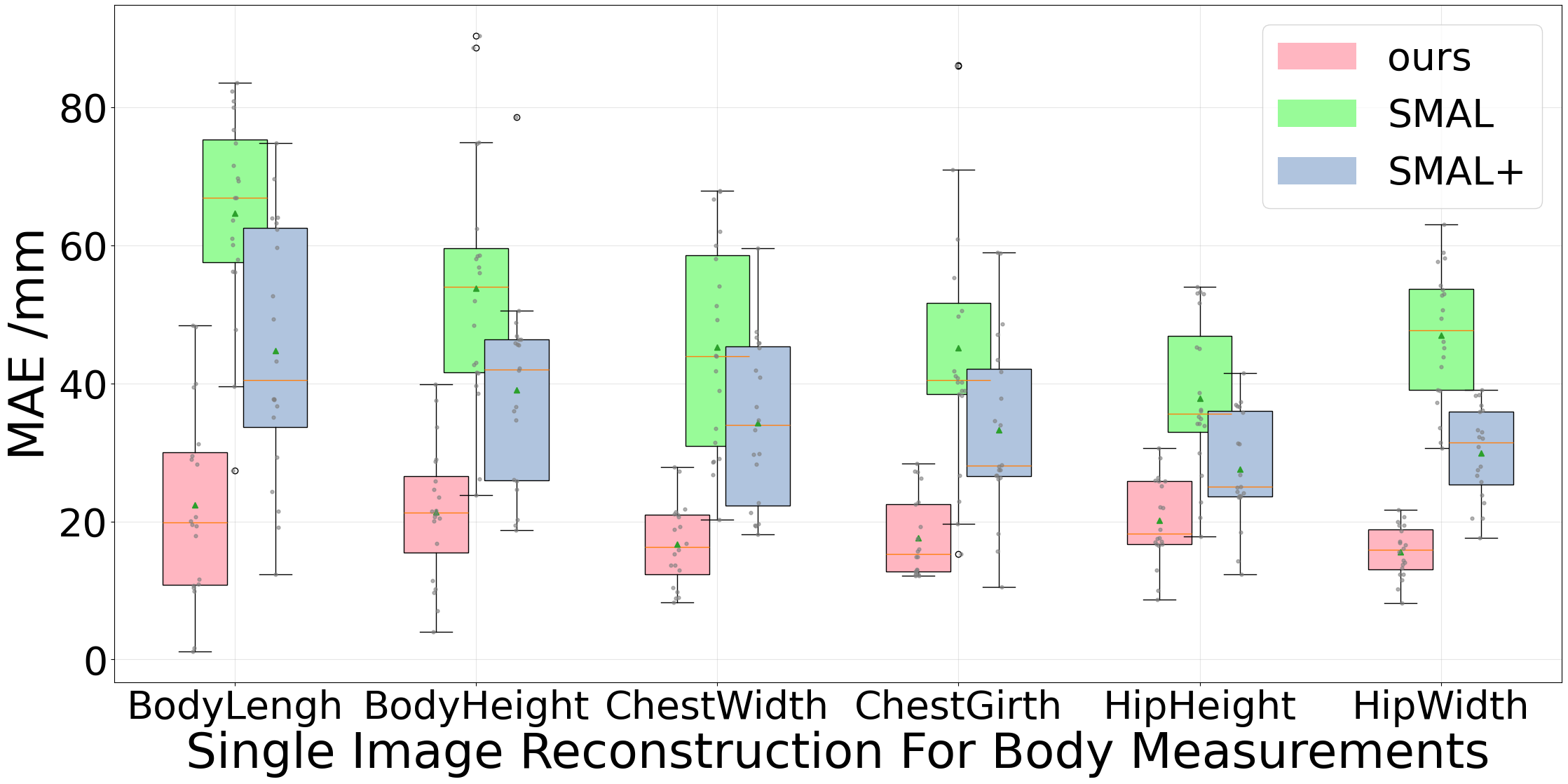}
    \caption{MAE results for body measurements: comparison between single-view SaanenGoat reconstructions and SMAL/SMAL+ baseline models. Errors in mm.}
    \label{fig:single_comparison_results}
\end{figure}

To validate our monocular estimation network, we test 20 Saanen dairy goats with available measurement data. Using MMPose for 2D keypoint detection, our method takes single-view RGBD images and optimizes pose/shape parameters through iterative refinement to reconstruct 3D meshes via the SaanenGoat parametric model. We compared against SMALify (SMAL, SMAL+) and AniMer, a family-aware Transformer approach for animal pose/shape estimation.
Fig.~\ref{fig:monocular_results} visualizes estimation results, while Table~\ref{tab:monocular_recovery_results_number} provides quantitative comparisons. SAANEN achieves 70.8\% and 65.9\% lower errors than SMAL and SMAL+, respectively. For body measurements from T-pose reconstructed meshes (Fig.~\ref{fig:single_comparison_results}), our method reduces average MAE by 42.2mm (20.6 vs 62.8mm) and 20.4mm (20.6 vs 41.0mm) compared to SMAL and SMAL+. MAPE results are in the supplementary material. These results demonstrate our network's reliability for single-view 3D reconstruction and body measurement estimation in Saanen dairy goats.

\begin{table}

\centering
\begin{tabular}{l cccc}
\toprule
& \multicolumn{2}{c}{Chamfer Distance} & \multicolumn{2}{c}{Mesh-to-Scan} \\
\cline{2-3} \cline{4-5}
& mean & std & Mean & std \\
\midrule
SMAL & 89.88 & 9.04 & 68.17 & 7.34 \\
SMAL+ & 76.16 & 8.34 & 59.32 & 6.07 \\
Ours & 26.58 & 3.17 & 19.66 & 1.74 \\
\bottomrule
\end{tabular}
\caption{Comparison of monocular recovery results between ours and SMAL as well as SMAL+ models. Errors in mm.}
\label{tab:monocular_recovery_results_number}
\end{table}

\section{Conclusion}
We address technical bottlenecks in 3D morphological data acquisition and analysis for precision livestock farming using female Saanen dairy goats. We establish the first authentic eight-view RGBD synchronized video dataset of dairy goats in real-world scenarios—FemaleSaanenGoat—containing 55 female Saanen specimens with diverse natural postures. Using multi-view Dynamic Fusion, we achieved high-fidelity 3D reconstruction and developed SaanenGoat, a specialized parametric model that decomposes goat morphology into interpretable pose and shape parameters. Based on anatomical landmarks, our system enables non-invasive rapid biometric measurement, automatically extracting six critical parameters: body length, withers height, shoulder breadth, thoracic circumference, pelvic width, and hip height.

Current limitations include species-specific constraints limiting generalization to other breeds and suboptimal reconstruction accuracy for occluded mammary regions. Future work will develop occlusion-aware reconstruction algorithms, expand datasets to multiple goat breeds, and address morphological challenges from varying wool thickness through multi-modal sensing.

Our framework provides a transferable foundation for 3D vision-based livestock analysis, advancing AI-driven agricultural applications by demonstrating how computer vision and parametric modeling address precision farming challenges while supporting livestock industry digitalization.

\section{Acknowledgments}
This work was supported by the National Key Research and Development Program of China (2022YFD1300200), the Xi'an Science and Technology Bureau Technology Tackling Special Plan Project (K4050724281),the Construction of “Scientist and Engineer” team of Qin Chuang Yuan in Shaanxi Province of China(2023KXJ-109), the National Natural Science Foundation of China (62125107), the Shuimu Tsinghua Scholar Program (2024SM324), and the Doctoral Graduate Student Independent Innovation Research Project of Northwest A\&F University (2025KYCXZ21).All data collection procedures were conducted in strict accordance with ethical standards, with no invasive interventions applied to the animals and no stress responses elicited.

\bibliography{aaai2026}

@String{Computer = "{IEEE} Computer" }

@String{Springer = "Springer-Verlag" }

@article{lu2025automatic,
  title={Automatic coarse-to-fine method for cattle body measurement based on improved GCN and 3D parametric model},
  author={Lu, Hexiao and Zhang, Jialong and Yuan, Xufeng and Lv, Jihong and Zeng, Zhiwei and Guo, Hao and Ruchay, Alexey},
  journal={Computers and Electronics in Agriculture},
  volume={231},
  pages={110017},
  year={2025},
  publisher={Elsevier}
}

@article{an2023three,
  title={Three-dimensional surface motion capture of multiple freely moving pigs using MAMMAL},
  author={An, Liang and Ren, Jilong and Yu, Tao and Hai, Tang and Jia, Yichang and Liu, Yebin},
  journal={Nature Communications},
  volume={14},
  number={1},
  pages={7727},
  year={2023},
  publisher={Nature Publishing Group UK London}
}

@inproceedings{biggs2020left,
  title={Who left the dogs out? 3d animal reconstruction with expectation maximization in the loop},
  author={Biggs, Benjamin and Boyne, Oliver and Charles, James and Fitzgibbon, Andrew and Cipolla, Roberto},
  booktitle={Computer Vision--ECCV 2020: 16th European Conference, Glasgow, UK, August 23--28, 2020, Proceedings, Part XI 16},
  pages={195--211},
  year={2020},
  organization={Springer},
}

@inproceedings{wu2023magicpony,
  title={Magicpony: Learning articulated 3d animals in the wild},
  author={Wu, Shangzhe and Li, Ruining and Jakab, Tomas and Rupprecht, Christian and Vedaldi, Andrea},
  booktitle={Proceedings of the IEEE/CVF Conference on Computer Vision and Pattern Recognition},
  pages={8792--8802},
  year={2023}
}

@inproceedings{xu2023animal3d,
  title={Animal3d: A comprehensive dataset of 3d animal pose and shape},
  author={Xu, Jiacong and Zhang, Yi and Peng, Jiawei and Ma, Wufei and Jesslen, Artur and Ji, Pengliang and Hu, Qixin and Zhang, Jiehua and Liu, Qihao and Wang, Jiahao and others},
  booktitle={Proceedings of the IEEE/CVF International Conference on Computer Vision},
  pages={9099--9109},
  year={2023}
}

@article{ruchay2020accurate,
  title={Accurate body measurement of live cattle using three depth cameras and non-rigid 3-D shape recovery},
  author={Ruchay, Alexey and Kober, Vitaly and Dorofeev, Konstantin and Kolpakov, Vladimir and Miroshnikov, Sergei},
  journal={Computers and Electronics in Agriculture},
  volume={179},
  pages={105821},
  year={2020},
  publisher={Elsevier}
}

@article{jin2024pointstack,
  title={PointStack based 3D automatic body measurement for goat phenotypic information acquisition},
  author={Jin, Bo and Wang, Guorui and Feng, Jingze and Qiao, Yongliang and Yao, Zhifeng and Li, Mei and Wang, Meili},
  journal={Biosystems Engineering},
  volume={248},
  pages={32--46},
  year={2024},
  publisher={Elsevier}
}

@article{du2022automatic,
  title={Automatic livestock body measurement based on keypoint detection with multiple depth cameras},
  author={Du, Ao and Guo, Hao and Lu, Jie and Su, Yang and Ma, Qin and Ruchay, Alexey and Marinello, Francesco and Pezzuolo, Andrea},
  journal={Computers and electronics in agriculture},
  volume={198},
  pages={107059},
  year={2022},
  publisher={Elsevier}
}

@article{hao2023improved,
  title={An improved PointNet++ point cloud segmentation model applied to automatic measurement method of pig body size},
  author={Hao, Hu and Jincheng, Yu and Ling, Yin and Gengyuan, Cai and Sumin, Zhang and Huan, Zhang},
  journal={Computers and electronics in agriculture},
  volume={205},
  pages={107560},
  year={2023},
  publisher={Elsevier}
}

@inproceedings{newcombe2015dynamicfusion,
  title={Dynamicfusion: Reconstruction and tracking of non-rigid scenes in real-time},
  author={Newcombe, Richard A and Fox, Dieter and Seitz, Steven M},
  booktitle={Proceedings of the IEEE conference on computer vision and pattern recognition},
  pages={343--352},
  year={2015}
}

@article{yang2023track,
  title={Track anything: Segment anything meets videos},
  author={Yang, Jinyu and Gao, Mingqi and Li, Zhe and Gao, Shang and Wang, Fangjing and Zheng, Feng},
  journal={arXiv preprint arXiv:2304.11968},
  year={2023}
}

@InProceedings{Zuffi_2017_CVPR,
author = {Zuffi, Silvia and Kanazawa, Angjoo and Jacobs, David W. and Black, Michael J.},
title = {3D Menagerie: Modeling the 3D Shape and Pose of Animals},
booktitle = {Proceedings of the IEEE Conference on Computer Vision and Pattern Recognition},
month = {July},
year = {2017}
}

@inproceedings{zuffi2018lions,
  title={Lions and tigers and bears: Capturing non-rigid, 3d, articulated shape from images},
  author={Zuffi, Silvia and Kanazawa, Angjoo and Black, Michael J},
  booktitle={Proceedings of the IEEE conference on Computer Vision and Pattern Recognition},
  pages={3955--3963},
  year={2018}
}

@article{li2021hsmal,
  title={hsmal: Detailed horse shape and pose reconstruction for motion pattern recognition},
  author={Li, Ci and Ghorbani, Nima and Broom{\'e}, Sofia and Rashid, Maheen and Black, Michael J and Hernlund, Elin and Kjellstr{\"o}m, Hedvig and Zuffi, Silvia},
  journal={arXiv preprint arXiv:2106.10102},
  year={2021}
}

@inproceedings{zuffi2024varen,
  title={VAREN: Very accurate and realistic equine network},
  author={Zuffi, Silvia and Mellbin, Ylva and Li, Ci and Hoeschle, Markus and Kjellstr{\"o}m, Hedvig and Polikovsky, Senya and Hernlund, Elin and Black, Michael J},
  booktitle={Proceedings of the IEEE/CVF Conference on Computer Vision and Pattern Recognition},
  pages={5374--5383},
  year={2024}
}

@inproceedings{rueegg2022barc,
  title={Barc: Learning to regress 3d dog shape from images by exploiting breed information},
  author={Rueegg, Nadine and Zuffi, Silvia and Schindler, Konrad and Black, Michael J},
  booktitle={Proceedings of the IEEE/CVF Conference on Computer Vision and Pattern Recognition},
  pages={3876--3884},
  year={2022}
}

@inproceedings{zuffi2024awol,
  title={AWOL: Analysis WithOut Synthesis Using Language},
  author={Zuffi, Silvia and Black, Michael J},
  booktitle={European Conference on Computer Vision},
  pages={1--19},
  year={2024},
  organization={Springer}
}

@inproceedings{li2024learning,
  title={Learning the 3d fauna of the web},
  author={Li, Zizhang and Litvak, Dor and Li, Ruining and Zhang, Yunzhi and Jakab, Tomas and Rupprecht, Christian and Wu, Shangzhe and Vedaldi, Andrea and Wu, Jiajun},
  booktitle={Proceedings of the IEEE/CVF Conference on Computer Vision and Pattern Recognition},
  pages={9752--9762},
  year={2024}
}

@inproceedings{zuffi2019three,
  title={Three-d safari: Learning to estimate zebra pose, shape, and texture from images" in the wild"},
  author={Zuffi, Silvia and Kanazawa, Angjoo and Berger-Wolf, Tanya and Black, Michael J},
  booktitle={Proceedings of the IEEE/CVF International Conference on Computer Vision},
  pages={5359--5368},
  year={2019}
}

@inproceedings{biggs2019creatures,
  title={Creatures great and smal: Recovering the shape and motion of animals from video},
  author={Biggs, Benjamin and Roddick, Thomas and Fitzgibbon, Andrew and Cipolla, Roberto},
  booktitle={Computer Vision--ACCV 2018: 14th Asian Conference on Computer Vision, Perth, Australia, December 2--6, 2018, Revised Selected Papers, Part V 14},
  pages={3--19},
  year={2019},
  organization={Springer}
}

@inproceedings{ruegg2023bite,
  title={BITE: Beyond priors for improved three-D dog pose estimation},
  author={R{\"u}egg, Nadine and Tripathi, Shashank and Schindler, Konrad and Black, Michael J and Zuffi, Silvia},
  booktitle={Proceedings of the IEEE/CVF Conference on Computer Vision and Pattern Recognition},
  pages={8867--8876},
  year={2023}
}

@article{niewiadomski2024generative,
  title={Generative Zoo},
  author={Niewiadomski, Tomasz and Yiannakidis, Anastasios and Cuevas-Velasquez, Hanz and Sanyal, Soubhik and Black, Michael J and Zuffi, Silvia and Kulits, Peter},
  journal={arXiv preprint arXiv:2412.08101},
  year={2024}
}

@inproceedings{lyu2025animer,
  title={AniMer: Animal Pose and Shape Estimation Using Family Aware Transformer},
  author={Lyu, Jin and Zhu, Tianyi and Gu, Yi and Lin, Li and Cheng, Pujin and Liu, Yebin and Tang, Xiaoying and An, Liang},
  booktitle={Proceedings of the Computer Vision and Pattern Recognition Conference},
  pages={17486--17496},
  year={2025}
}

@inproceedings{zhang1999flexible,
  title={Flexible camera calibration by viewing a plane from unknown orientations},
  author={Zhang, Zhengyou},
  booktitle={Proceedings of the seventh ieee international conference on computer vision},
  volume={1},
  pages={666--673},
  year={1999},
  organization={Ieee}
}

@inproceedings{kearney2020rgbd,
  title={Rgbd-dog: Predicting canine pose from rgbd sensors},
  author={Kearney, Sinead and Li, Wenbin and Parsons, Martin and Kim, Kwang In and Cosker, Darren},
  booktitle={Proceedings of the IEEE/CVF conference on computer vision and pattern recognition},
  pages={8336--8345},
  year={2020}
}

@article{yao2022lassie,
  title={Lassie: Learning articulated shapes from sparse image ensemble via 3d part discovery},
  author={Yao, Chun-Han and Hung, Wei-Chih and Li, Yuanzhen and Rubinstein, Michael and Yang, Ming-Hsuan and Jampani, Varun},
  journal={Advances in Neural Information Processing Systems},
  volume={35},
  pages={15296--15308},
  year={2022}
}

@article{hong2023lrm,
  title={Lrm: Large reconstruction model for single image to 3d},
  author={Hong, Yicong and Zhang, Kai and Gu, Jiuxiang and Bi, Sai and Zhou, Yang and Liu, Difan and Liu, Feng and Sunkavalli, Kalyan and Bui, Trung and Tan, Hao},
  journal={arXiv preprint arXiv:2311.04400},
  year={2023}
}

@article{sun2023dreamcraft3d,
  title={Dreamcraft3d: Hierarchical 3d generation with bootstrapped diffusion prior},
  author={Sun, Jingxiang and Zhang, Bo and Shao, Ruizhi and Wang, Lizhen and Liu, Wen and Xie, Zhenda and Liu, Yebin},
  journal={arXiv preprint arXiv:2310.16818},
  year={2023}
}

@article{tochilkin2024triposr,
  title={Triposr: Fast 3d object reconstruction from a single image},
  author={Tochilkin, Dmitry and Pankratz, David and Liu, Zexiang and Huang, Zixuan and Letts, Adam and Li, Yangguang and Liang, Ding and Laforte, Christian and Jampani, Varun and Cao, Yan-Pei},
  journal={arXiv preprint arXiv:2403.02151},
  year={2024}
}

@article{zhang2025whole,
  title={Whole-genome variants resource of 298 Saanen dairy goats},
  author={Zhang, Kai and Zhao, Jianqing and Mi, Shirong and Liu, Jiqiang and Luo, Jun and Liu, Jianxin and Shi, Hengbo},
  journal={Scientific Data},
  volume={12},
  number={1},
  pages={528},
  year={2025},
  publisher={Nature Publishing Group UK London}
}

@article{wang2022genome,
  title={Genome-wide association analysis of milk production, somatic cell score, and body conformation traits in Holstein cows},
  author={Wang, Peng and Li, Xue and Zhu, Yihao and Wei, Jiani and Zhang, Chaoxin and Kong, Qingfang and Nie, Xu and Zhang, Qi and Wang, Zhipeng},
  journal={Frontiers in Veterinary Science},
  volume={9},
  pages={932034},
  year={2022},
  publisher={Frontiers Media SA}
}

@incollection{loper2023smpl,
  title={SMPL: A skinned multi-person linear model},
  author={Loper, Matthew and Mahmood, Naureen and Romero, Javier and Pons-Moll, Gerard and Black, Michael J},
  booktitle={Seminal Graphics Papers: Pushing the Boundaries, Volume 2},
  pages={851--866},
  year={2023}
}

@article{lyu2025animer+,
  title={AniMer+: Unified Pose and Shape Estimation Across Mammalia and Aves via Family-Aware Transformer},
  author={Lyu, Jin and An, Liang and Lin, Li and Cheng, Pujin and Liu, Yebin and Tang, Xiaoying},
  journal={arXiv preprint arXiv:2508.00298},
  year={2025}
}


\end{document}